\documentclass[times,twocolumn,final]{elsarticle}

\usepackage{lineno,hyperref}

\journal{Journal of \LaTeX\ Templates}

\usepackage{framed,multirow}
\usepackage{adjustbox}
\usepackage{amssymb}
\usepackage{latexsym}
\usepackage[margin=0.7in]{geometry}
\usepackage{multirow}
\usepackage{tabularx}

\usepackage{algorithm}
\usepackage{algorithmicx}
\usepackage{etoolbox}
\usepackage{algpseudocode}
\usepackage{times}
\usepackage{epsfig}
\usepackage{amssymb}
\usepackage{bbm}
\usepackage{xcolor}
\usepackage{amsmath} 

\usepackage{mathtools}
\usepackage{soul}
\usepackage{xcolor}         
\usepackage{bm}             
\usepackage[capitalise]{cleveref}       
\usepackage{graphicx}       
\usepackage{booktabs}       
\usepackage{amsfonts}       
\usepackage{nicefrac}       
\usepackage{microtype}      
\usepackage{subcaption}

\mathchardef\mhyphen="2D 
  
\definecolor{red}{rgb}{0.234, 0.6835, 0.808}
\usepackage{color, colortbl}
\definecolor{Gray}{gray}{0.9}

\begin{document}


\begin{frontmatter}

\title{Adaptive Contrastive Learning with Dynamic Correlation for Multi-Phase Organ Segmentation}

\author{Ho Hin Lee\textsuperscript{1}, Yucheng Tang\textsuperscript{2}, Han Liu\textsuperscript{1}, Yubo Fan\textsuperscript{1}, Leon Y. Cai\textsuperscript{3}, Qi Yang\textsuperscript{1}, Xin Yu\textsuperscript{1}, Shunxing Bao\textsuperscript{1}, Yuankai Huo\textsuperscript{1,2}, Bennett A. Landman\textsuperscript{1,2,4}}

\address{\textsuperscript{1}Department of Computer Science, Vanderbilt University, Nashville, TN, USA 37212}
\address{\textsuperscript{2}Department of Electrical and Computer Engineering, Vanderbilt University, Nashville, TN, USA, 37212}
\address{\textsuperscript{3}Department of Biomedical Engineering, Vanderbilt University, Nashville, TN, USA, 37212}
\address{\textsuperscript{4}Radiology, Vanderbilt University Medical Center, Nashville, TN, USA, 37235}


\begin{abstract}
Recent studies have demonstrated the superior performance of introducing ``scan-wise" contrast labels into contrastive learning for multi-organ segmentation on multi-phase computed tomography (CT). However, such scan-wise labels are limited: (1) a coarse classification, which could not capture the fine-grained ``organ-wise" contrast variations across all organs; (2) the label (i.e., contrast phase) is typically manually provided, which is error-prone and may introduce manual biases of defining phases. In this paper, we propose a novel data-driven contrastive loss function that adapts the similar/dissimilar contrast relationship between samples in each minibatch at organ-level. Specifically, as variable levels of contrast exist between organs, we hypothesis that the contrast differences in the organ-level can bring additional context for defining representations in the latent space. An organ-wise contrast correlation matrix is computed with mean organ intensities under one-hot attention maps. The goal of adapting the organ-driven correlation matrix is to model variable levels of feature separability at different phases. We evaluate our proposed approach on multi-organ segmentation with both non-contrast CT (NCCT) datasets and the MICCAI 2015 BTCV Challenge contrast-enhance CT (CECT) datasets. Compared to the state-of-the-art approaches, our proposed contrastive loss yields a substantial and significant improvement of 1.41\% (from 0.923 to 0.936, p-value$<$0.01) and 2.02\% (from 0.891 to 0.910, p-value$<$0.01) on mean Dice scores across all organs with respect to NCCT and CECT cohorts. We further assess the trained model performance with the MICCAI 2021 FLARE Challenge CECT datasets and achieve a substantial improvement of mean Dice score from 0.927 to 0.934 (p-value$<$0.01). The code is available at: \url{https://github.com/MASILab/DCC_CL}
\end{abstract}


\end{frontmatter}



\section{Introduction}

\begin{figure*}
\centering
\includegraphics[width=\textwidth]{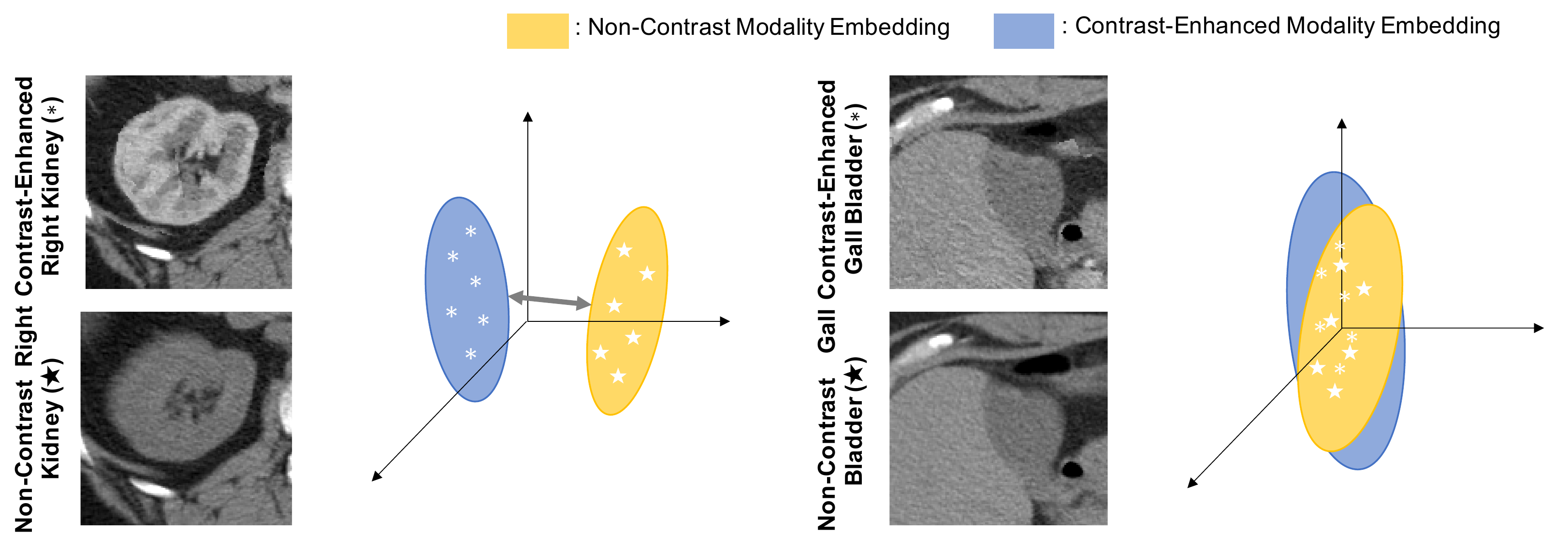}
\caption{\footnotesize{Between NCCT and CECT, some organs have significant contrast variation (such as the kidneys, liver, and spleen). We expect the corresponding embedding to be separable by contrast phases. However, some organs of interest such as the gall bladder, pancreas, and adrenal glands have similar contrast appearance across both modalities. As such, we expect the embedding of these organs to be aligned across phases, instead of separating into independent clusters using contrast label supervision. Such variation exposes a key limitation of current contrastive learning state-of-the-art approaches.}} \label{problem_figure}
\end{figure*}

\label{sec1}

Multi-phase contrast CT delineates different anatomical structures with significant variation in contrast intensity \cite{tang2021phase, heiken1995dynamic}. As contrast agents progress through blood vessels, each organ of interest has its distinctive contrast uptake patterns, which leads to a wide range of intensities. It is challenging to train a generalizable deep learning model to achieve consistent segmentation performance across such large variations. The most naive way to adapt multi-phase representations is to train multiple networks to compute and fuse phase-specific features for organs segmentation \cite{raju2021coheterogeneous}. Phase/Modality-dependent normalization is further proposed to separate normalization layers to specific imaging phases and mitigate the discrepancy between phases \cite{dou2020unpaired}. However, limited studies have been proposed to study the multi-phase correspondence of abdominal organs in the feature level. Contrastive learning, a variant of self-supervised learning, has been shown to learn class-aware representations in a ``scan-wise" setting and achieves significant improvements in the downstream tasks \cite{chen2020simple, tian2019contrastive, chaitanya2020contrastive, wang2021dense}. The theory of contrastive learning consists of two main operations: 1) pull the representation of target image (anchor) and a matching sample together as a ``positive pair", and 2) push the anchor representation from the remaining non-matching samples apart as ``negative pairs". The goal of contrastive learning is to define class-aware embeddings without additional guidance. Furthermore, supervised contrastive learning is introduced to further enhance the ability of defining class-wise embeddings with the scan-wise label given \cite{khosla2020supervised, hin2021semantic}. However, scan-wise contrast information (e.g., phases) is limited in representing different scales of variations in contrast intensity across organs of interest. For example, we observed that kidney organ demonstrates a distinctive comparison in appearances between contrast-enhanced and non-contrast phase, but only subtle variance is demonstrated in appearance of gall bladder (in Fig.1). Therefore, it is challenging to leverage single hard label to represent the dynamic changes across organ intensities. As such, we can further ask: \textbf{Can we leverage the appearance difference to control and define organ-wise representations with contrastive learning?}

\indent In this work, we propose a data-driven contrastive loss that leverages contrast correspondences between organs to adaptively model the representations into organ-specific embeddings. Specifically, we extract the mean intensity under the organ-specific regions with the one-hot attention guidance and compare the pairwise contrast similarity/dissimilarity across all samples in each minibatch. The goal of using contrast correlation is to dynamically regularize the contrastive loss based on the wide range of contrast enhancement. Our proposed contrastive loss is evaluated with two public contrast-enhanced CT (CECT) cohorts and one research non-contrast CT (NCCT) cohort. The experimental results demonstrate consistent improvement in multi-organ segmentation using deeplabv3+ architectures with a ResNet-50 encoder backbone \cite{he2016deep, chen2018encoder}. Our main contributions are summarized as follows:
\begin{itemize}
    \item We propose an adaptive contrastive learning framework to improve and generalize multi-organ segmentation performance from multi-phase contrast CT.
    \item We propose a data-driven contrastive loss function that adapts the organ-specific contrast correlation as an adaptive weighting constraint to control the distance between multi-phase representations on the organ-level.
    \item We demonstrate that the proposed contrastive loss captures the wide range of multi-phase contrast differences without trading off performance from one of the phases and achieves significant improvement for downstream segmentation tasks.
\end{itemize}

\begin{figure*}
\centering
\includegraphics[width=0.85\textwidth]{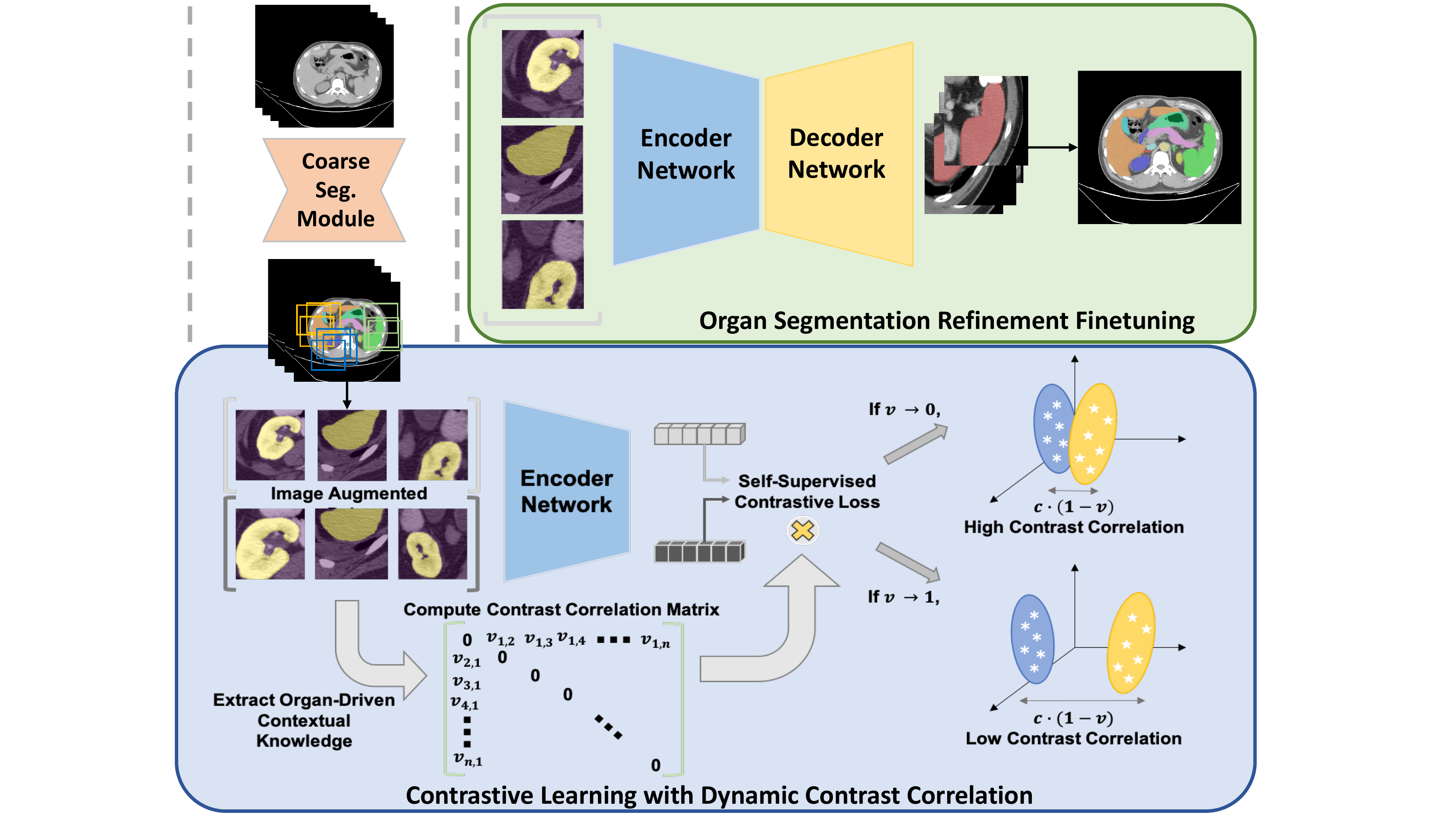}
\caption{\footnotesize{The complete contrastive framework can be divided into three hierarchical steps: 1) We first compute coarse segmentation and extract organ-corresponding patches for contrastive learning. The organ-specific attention masks are leveraged as additional channels to guide the representations extracted within specific regions. 2) For contrastive learning, we compute a contrast correlation matrix between samples in each minibatch to control the weighting of the contrastive loss dynamically across all pairs. We first compute the mean intensity of each organ of interest under each one-hot attention bounded region. \textbf{v} is the mean intensity difference between each augmented sample and \textbf{n} is the batch size of each minibatch. If the value of \textbf{v} is subtle, it corresponds to high contrast correlation between modalities and the cosine distance between representations \textbf{c} approaches 1. 3) We finally finetune the well-pretrained encoder followed with a decoder network to generate refine multi-organ segmentation.}} \label{pipeline_figure}
\end{figure*}

\section{Related Works}

\textbf{Medical Image Segmentation:} 
Most modern approaches to perform medical image segmentation typically train a deep neural network directly in supervised setting with post-processing techniques \cite{wang2021view}. However, the model performance is greatly dependent on both the quality of ground-truth labels and the resolution of volumes \cite{gamechi2019automated}. Patch-wise approaches and hierarchical approaches are proposed to adapt coarse-to-fine features and leverage the mutli-scale capabilities to generate refine segmentation \cite{roth2018multi, zhu2019multi, tang2021high}. However, multiple models are typically needed to train for multiple semantic targets segmentation. Single coarse-to-refine network is proposed to adapt multi-organ segmentation by integrating binary organ-corresponding attentions as additional input channel \cite{lee2021rap}. To further enhance the segmentation performance with stability, significant efforts are put into exploring the possibility of adapting unlabeled data in both semi-supervised or self-supervised setting. Generating quality assurance score for predicted mask are proposed as an alternative supervision using unlabeled data \cite{lee2020semi}. Different pretext tasks including colorization, deformation and image rotation, have been leveraged as pretraining strategies to provide a better initialization for segmentation networks \cite{zhou2021models}. Furthermore, learning spatial context by predicting the degree of rotation and relative patch position are proposed to be beneficial for finetuning segmentation network \cite{bai2019self, zhuang2019self}.

\textbf{Image-Level Contrastive Learning:} Significant efforts have been put into self-supervised learning to extract meaningful representations from unlabeled data. Previous works have demonstrated to learn representations in the latent space using image reconstruction \cite{zhang2017split}. The idea of contrastive learning further extends to learn and classify representations into embeddings by evaluating the pairwise similarity in the latent space. By leveraging data augmentations, the augmented pairs from the same image are defined as the positive pair and pull their representations closer together, while pushing the remaining representations apart as negative pairs \cite{chen2020simple}. Instead of evaluating the feature similarity, maximizing the mutual information between representations have also been proposed to correlate the similar representations in the latent space \cite{bachman2019learning}. As the feature similarity is evaluated within a batch, increasing the batch size is another alternative to enhance the efficiency of contrastive learning. Using memory bank and momentum encoder for contrastive learning have been proposed to compare the query representations with more dissimilar representations within a minibatch \cite{misra2020self, he2020momentum}. Another perspective of contrastive learning is the definition of positive pair. Representations from the same class may also sample in the minibatch, while the contrastive loss can only select a specific positive pair. Supervised contrastive learning is introduced and leverage the image-level label to define arbitrary number of positive pairs and provide a better definition of the latent space for fintuning classification tasks \cite{khosla2020supervised}. In the medical domain, continuous proxy meta-data is leveraged as the image-level additional guidance for contrastive learning with brain MRI \cite{dufumier2021contrastive}. Furthermore, leveraging relative position of the extracted patch for contrastive learning is demonstrated to be beneficial for fintuning detection task \cite{lei2021contrastive}. For medical image segmentation, image-wise labels such as positional information, are additionally use to constrain representations in the latent space \cite{zeng2021positional}. To tackle the multi-contrast phase organ segmentation, the contrastive loss have extended to adapt multi-class "scan-wise" labels (e.g. contrast phases and organs) and define representations into sub-classes embeddings to benefit segmentation performance. However, such hard labels are difficult to represent the dynamic contrast level across organ of interests. Current contrastive approaches are limited to provide flexibility of controlling the separability between representations adaptively.

\textbf{Pixel-Level Contrastive Learning:}
While prior works have demonstrated leveraging contrastive learning to enhance "image-wise" downstream task performance, several approaches have been extended the theory basis of contrastive learning to pixel-wise   setting. Instead of using linear projection, dense projection is used to compute dense mapping and evaluate the pixel-level similarity in self-supervised setting \cite{wang2021dense}. After that, pixel-wise contrastive loss is proposed to evaluate the feature similarity with ground-truth label guidance \cite{zhao2021contrastive}. Apart from leveraging contrastive learning as pretraining strategies, single-stage framework is proposed to cotrain segmentation task with contrastive loss and enforces the representations in the same semantic class to be more similar as independent pixel-wise embeddings \cite{wang2021exploring, hu2021region, alonso2021semi}. In the medical perspective, feature representations in both local and global view are computed, and evaluate the structural similarity across views with limited samples \cite{chaitanya2020contrastive}. Furthermore, limited number of ground truth-labels or pseudo predicted labels are further adapted with the pixel-wise contrastive loss and enhance the downstream segmentation performances \cite{hu2021semi, liu2022margin}. However, such pretraining/co-training strategies still require pixel-wise label guidance.

\begin{figure*}
\centering
\includegraphics[width=\textwidth]{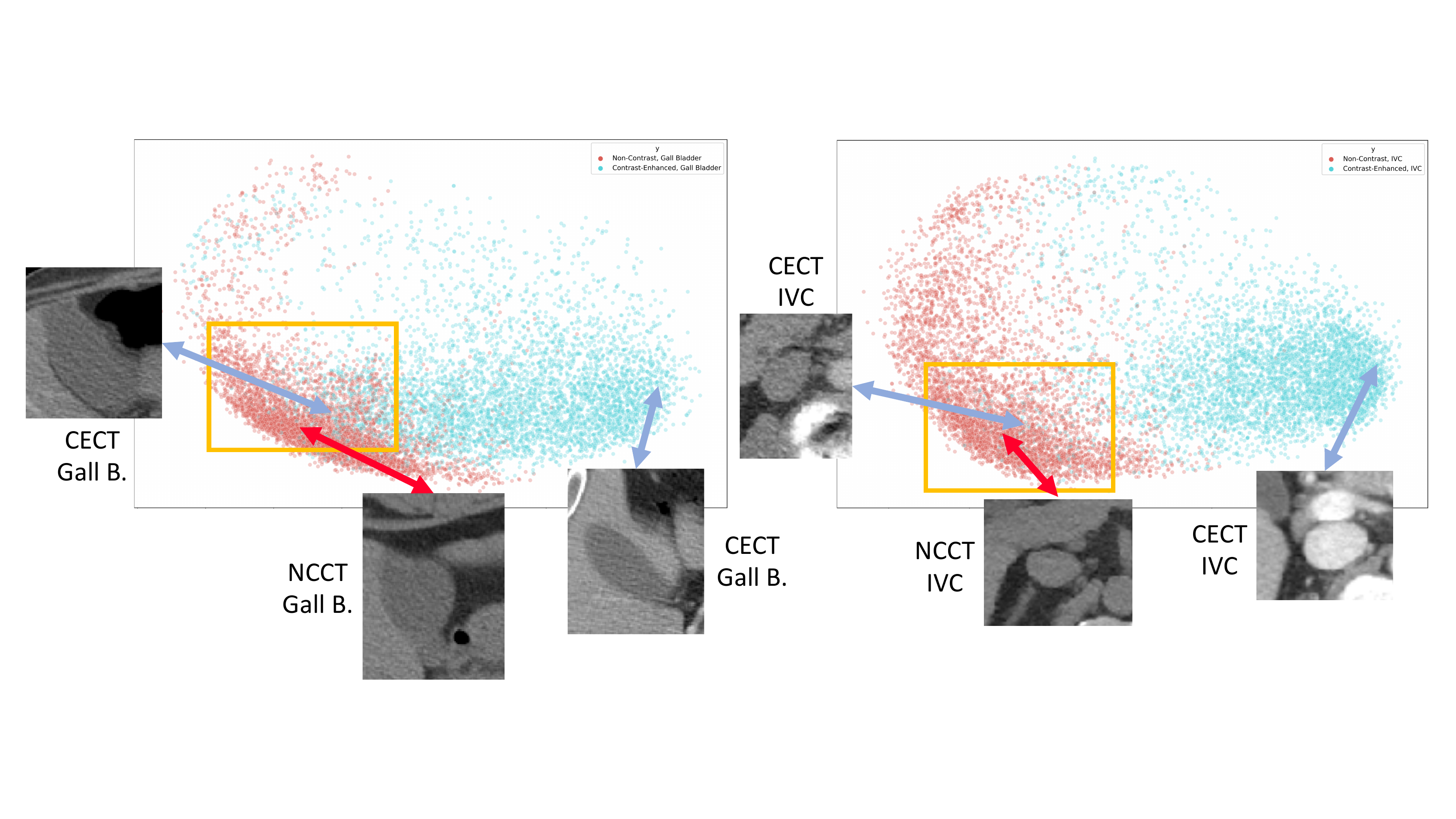}
\caption{{Dimensionality reduction with PCA is performed to visualize the distribution of learned representations. Both left and right plots are corresponding to the gall bladder feature and IVC feature respectively. By leveraging the contrast correlation as dynamic weighting, we found that the representations is well defined according to the contrast level, although they are in different "scan-wise" defined contrast phases.}} \label{pca_figure}
\end{figure*}

\section{Methods}
The overall goal of our proposed method is to dynamically extract contrast-related knowledge from organs themselves across contrast phases and define a robust multi-phase latent space for the organ segmentation task. The complete hierarchical pipeline is presented in Fig. 2. We can divide our hierarchical pipeline into three specific stages: 1) extracting organ-specific attention from coarse segmentation, 2) contrastive learning with contrast correlation, and 3) fine-tuning for organ segmentation refinement.


\subsection{Organ-Specific Attention from Coarse Segmentation}
Given a set of multi-contrast 3D image volumes $V_i = \{X_i, Y_i\}_{i=1,...,L}$, where $L$ is the number of all imaging samples, $X$ is the volumetric image and $Y$ is the corresponding multi-organ label. Inspired by \cite{hin2021semantic}, we leverage a low-resolution (complete volume) segmentation network $Coarse(\cdot)$ to generate coarse segmentation masks $A_i=Coarse(X_i)$ for multiple organs. We define $A\in{R^{H\times W\times C}}$, where $H$ and $W$ denote as the axial dimension of the image, and $C$ denotes as the total number of organ label classes. The coarse segmentation network is implemented with a hierarchical 3D U-Net architecture and is trained by the complete multi-contrast volumes with 5-fold cross-validation \cite{lee2021rap}. With the voxel-wise guidance from multi-organ masks, we randomly sample organ-specific patches for contrastive pretraining and downstream task fine-tuning. Briefly, organ-specific patches $p_i=\{x_{C,i}, y_{C,i}, s_{C,i}\}_{i=1,...,N}$ are extracted by randomly sampling voxel index of each organ class $C$ as the center point to crop the regions of interest (ROIs), $N$ denotes as the total number of query patches, $x_{C,i}$ is the image patch. Furthermore, we extract $y_{C,i}$ and $s_{C,i}$ by converting the organ ROIs in both ground-truth label patch and the coarse segmentation patch in binary setting, where $s_{C,i}\in {A_i}$. The coarse mapping is  utilized as the organ-specific attention map for spatial restrictions of learning voxel-wise semantic representations. Specifically, inspired by \cite{chen2020simple}, a data augmentation module $Aug(\cdot)$ is used to further extract correlated representations within the organ ROIs, including random cropping, rotation (-30 to 30 degrees), scaling (width: 0.3, height: 0.7) and generate $2n$ pairwise copies $\widetilde{x}_{C,i},\widetilde{s}_{C,i} = Aug({x_{C,i}, s_{C,i}})$. Each $\widetilde{s}_{C,i}$ is concatenated with $\widetilde{x}_{C,i}$ as a multi-channel input $m_i$ for training encoder network $E(\cdot)$.


\subsection{Contrastive Learning with Contrast Correlation}
\subsubsection{Phase-Driven Contrast Correlation}
From Fig. 1, we observe that the contrast uptake patterns between organs of interest are varied significantly and subtle variations also exist in some of the organs, even when they are in different modalities. Therefore, we hypothesize that the dynamic intensity  difference in organ level is beneficial to define representations in the latent space. Here, we extract the mean intensity value $d_i$ within the corresponding $\widetilde{s}_{C,i}$ and dynamically compute the contrast difference pairwisely across all image patches of each minibatch. The absolute difference of mean intensity across pairwise samples $v_{i,j}$ is computed as the contrast correlation context. We hypothesize that $v_{i,j}$ tends to 0 if the ``organ-driven" mean intensity is similar. The correlation matrix is defined as following:
\begin{equation}
d_{i} = \frac{1}{|\phi|} \sum_{{(x,y,c)}\in{\phi}}{\{\widetilde{x}_{C,i}\cdot\widetilde{s}_{C,i}\}}(x,y,c)
\end{equation}
\begin{equation}
v_{i,j} = |d_{i} - d_{j}|\:,
\end{equation}
where $x$, $y$ and $c$ are corresponding to the x-y coordinates and number of channels within the attention bounded region. $\phi\in R$ is the number of nonzero pixels in the bounded region. $i$ and $j$ are the respective batch indices across all augmented samples. 

\subsubsection{Dynamic Contrast Correlation Contrastive Loss}
Previously, \textit{Lee et al.} used image-wise modality labels to constrain the pairwise representation into corresponding modality embeddings \cite{hin2021semantic}. However, image-wise labels cannot account the accurate variation in contrast enhancement across different organs. Thus, current contrastive loss functions lack of flexibility in adjusting the level of separation dynamically between the multi-phase/modal representations. Therefore, we propose the dynamic contrast correlation (DCC) matrix as an adaptive weighting to multiply the cosine similarity computed in the feature-level. Instead of only computing feature-level contrastive loss in self-supervised setting, the dynamic constraints account the image-level variability as soft supervision to enhance the flexibility of standard contrastive loss, which help to control the cosine distance between the pairwise representations in each minibatch. The DCC matrix is dynamically varies according to the shuffled samples in each minibatch. We define our proposed contrastive loss as $\mathcal{L}_{dcc}$ and is defined respectively as following:

\begin{equation}
    \widetilde{z}_{k}, \widetilde{z}_{p(k)} = P(E(m_{2k}, m_{2k-1})
\end{equation}

\begin{equation}
    \mathcal{L}_{dcc} = -\sum_{k=1}^{2n} \log \frac{{\exp(\widetilde{z}_{k} \cdot \widetilde{z}_{p(k)} \cdot (1-v_{k,p(k)})/\mathcal{T})}}{{\sum_{j\in{J(k)}}} \exp({\widetilde{z}_{k} \cdot \widetilde{z}_{j} \cdot (1-v_{k,j})/\mathcal{T})}}\:,
\end{equation}
where  $\widetilde{z}_{k}$ and $\widetilde{z}_{p(k)}$ are the pairwise feature representation vectors. The index $k$ represents the sample of anchor and index $p(k)$ represents the corresponding positive. The hyperparameter $\mathcal{T}$ is the radius of the hypersphere that maps the representation inside as a point. $P(\cdot)$ is a linear projection network using multi-layer perceptron (MLP). Instead of constraining into label-class representation, the DCC-weighted contrastive loss preserve the data-driven knowledge for each organ and allows for similar organ-wise representations even if they are from different phases. The distance between the pairwise representations is controlled with the contrast correlation $(1-v_{k,j})$ and enhance the flexibility of defining latent space.

\newcommand{\tabincell}[2]{\begin{tabular}{@{}#1@{}}#2\end{tabular}}
\begin{table*}[htb]
    \centering
    \caption{Comparison of current state-of-the-art methods on the BTCV challenge leaderboard. (We only show Dice scores of 8 main organs due to limited space, $\star$: fully-supervised approach,  $\bigtriangleup$: partially supervised approach, *: $p<0.01$, with Wilcoxon signed-rank test.)}
    \begin{tabular}{*{1}{l}|*{8}{c}|*{2}{c}}
        \toprule 
         Method & Spleen & R.Kid & L.Kid & Gall. & Eso. & Liver & Aorta & IVC &\tabincell{c}{Average\\ Dice} \\
        
         \midrule
         \cite{cciccek20163d}$\star$ & 0.906 & 0.857 & 0.899 & 0.644 & 0.684 & 0.937 & 0.886 & 0.808 & 0.784 \\
         \cite{roth2018multi}$\star$ & 0.935 & 0.887 & 0.944 & 0.780 & 0.712 & 0.953 & 0.880 & 0.804 & 0.816 \\
         \cite{heinrich2015multi} & 0.920 & 0.894 & 0.915 & 0.604 & 0.692 & 0.948 & 0.857 & 0.828 & 0.790 \\
         \cite{pawlowski2017dltk}$\star$ & 0.939 & 0.895 & 0.915 & 0.711 & 0.743 & 0.962 & 0.891 & 0.826 & 0.815 \\
         \cite{zhu2019multi}$\star$ & 0.935 & 0.886 & 0.944 & 0.764 & 0.714 & 0.942 & 0.879 & 0.803 & 0.814 \\
         \cite{lee2021rap}$\star$ & 0.959 & 0.920 & 0.945 & 0.768 & 0.783 & 0.962 & 0.910 & 0.847 & 0.842 \\
         \cite{isensee2021nnu}$\star$ & 0.956 & 0.923 & 0.940 & 0.760 & 0.764 & 0.965 & 0.905 & 0.850 & 0.839 \\
         \cite{hatamizadeh2022unetr}$\star$ & 0.959 & 0.912 & 0.940 & 0.724 & 0.746 & 0.968 & 0.905 & 0.840 & 0.836\\
         \cite{Zhou_2019_ICCV}$\bigtriangleup$ & 0.968 & 0.920 & 0.953 & 0.729 & 0.790 & 0.974 & 0.925 & 0.847 & 0.850 \\
         \midrule
         \cite{chen2020simple} & 0.953 & 0.922 & 0.930 & 0.830 & 0.822 & 0.972 & 0.899 & 0.874 & 0.863 \\
         \cite{chaitanya2020contrastive} & 0.956 & 0.935 & 0.946 & 0.920 & 0.854 & 0.970 & 0.915 & 0.893 & 0.874 \\
         \cite{wang2021dense} & 0.963 & 0.939 & 0.900 & 0.815 & 0.838 & 0.976 & 0.922 & 0.907 & 0.882 \\
         \cite{khosla2020supervised} & 0.959 & 0.939 & 0.947 & \textbf{0.932} & 0.867 & 0.978 & 0.922 & 0.911 & 0.907 \\
         \cite{alonso2021semi} & 0.954 & 0.933 & 0.932 & 0.903 & 0.858 & 0.973 & 0.918 & 0.904 & 0.890 \\
         \cite{wang2021exploring} & 0.966 & 0.942 & 0.955 & 0.886 & 0.860 & 0.975 & 0.930 & 0.908 & 0.913\\
         \cite{hin2021semantic} & 0.971 & 0.955 & \textbf{0.963} & 0.910 & 0.886 & 0.984 & 0.941 & 0.932 & 0.923 \\ 
         \midrule
         \textbf{Ours (DCC-CL)} & \textbf{0.974} & \textbf{0.956} & 0.960 & 0.928 & \textbf{0.905} & \textbf{0.986} & \textbf{0.950} & \textbf{0.939} & \textbf{0.936*} \\
         \bottomrule
    \end{tabular}
    \label{baselines_compare}
\end{table*}

\begin{table*}[!htb]
    \centering
    \caption{Comparison of the current state-of-the-art methods on the non-contrast testing dataset. (We only show Dice scores of 8 main organs due to limited space, $\star$: fully-supervised approach, $\bigtriangleup$: partially supervised approach, *: $p<0.01$, with Wilcoxon signed-rank test.)}
    \begin{tabular}{*{1}{l}|*{8}{c}|*{3}{c}}
        \toprule 
         Method & Spleen & R.Kid & L.Kid & Gall. & Eso. & Liver & Aorta & IVC &\tabincell{c}{Average\\ Dice} \\
         \midrule
         \cite{cciccek20163d}$\star$ & 0.937 & 0.856 & 0.912 & 0.690 & 0.631 & 0.920 & 0.880 & 0.769 & 0.762 \\
         \cite{roth2018multi}$\star$ & 0.940 & 0.890 & 0.923 & 0.701 & 0.724 & 0.948 & 0.878 & 0.770 & 0.771 \\
         \cite{heinrich2015multi} & 0.910 & 0.865 & 0.889 & 0.624 & 0.656 & 0.930 & 0.860 & 0.759 & 0.748 \\
         \cite{zhu2019multi}$\star$ & 0.948 & 0.880 & 0.920 & 0.710 & 0.734 & 0.950 & 0.879 & 0.803 & 0.790 \\
         \cite{lee2021rap}$\star$ & 0.954 & 0.874 & 0.928 & 0.701 & 0.753 & 0.958 & 0.897 & 0.794 & 0.798 \\
         \cite{isensee2021nnu}$\star$ & 0.963 & 0.930 & 0.946 & 0.876 & 0.792 & 0.970 & 0.919 & 0.843 & 0.836 \\
         \cite{hatamizadeh2022unetr}$\star$ & 0.971 & 0.923 & 0.947 & 0.857 & 0.789 & 0.965 & 0.908 & 0.821 & 0.815\\
         \cite{Zhou_2019_ICCV}$\bigtriangleup$ & 0.960 & 0.900 & 0.943 & 0.739 & 0.810 & 0.965 & 0.920 & 0.810 & 0.833 \\
         \midrule
         \cite{chen2020simple} & 0.957 & 0.932 & 0.937 & 0.800 & 0.801 & 0.969 & 0.901 & 0.869 & 0.848 \\
         \cite{chaitanya2020contrastive} & 0.969 & 0.940 & 0.955 & 0.910 & 0.834 & 0.970 & 0.911 & 0.867 & 0.854 \\
         \cite{alonso2021semi} & 0.965 & 0.945 & 0.957 & 0.914 & 0.837 & 0.974 & 0.918 & 0.910 & 0.873 \\
         \cite{wang2021dense} & 0.979 & 0.961 & 0.964 & 0.941 & 0.865 & 0.979 & 0.937 & 0.923 & 0.887 \\
         \cite{khosla2020supervised} & 0.975 & 0.952 & 0.962 & 0.943 & 0.857 & 0.976 & 0.925 & 0.915 & 0.879 \\
         \cite{wang2021exploring} & 0.970 & 0.960 & 0.949 & 0.940 & 0.832 & 0.974 & 0.922 & 0.920 & 0.872 \\
         \cite{hin2021semantic} & 0.982 & 0.962 & 0.965 & 0.834 & 0.879 & 0.982 & 0.945 & 0.929 & 0.892 \\
         \midrule
         \textbf{Ours (DCC-CL)} & \textbf{0.984} & \textbf{0.966} & \textbf{0.970} & \textbf{0.957} & \textbf{0.890} & \textbf{0.982} & \textbf{0.951} & \textbf{0.939} & \textbf{0.910*} \\
         \bottomrule
    \end{tabular}
    \label{tab:my_label}
\end{table*}

\subsection{Fine-tuning for Organ Segmentation Refinement}
The final goal of our proposed approach is to learn the contrast correlated semantic representations for segmentation refinement. After the encoder network is pretrained with a projection network, we withdraw the projection network and employ DeepLabV3+ network as the model backbone for segmentation refinement. Such backbone can share the same encoder structure with the contrastive pretrained encoder and adapts atrous spatial pyramid pooling (ASPP) as the decoder for multi-scale refinement. Here, we use the Dice loss to evaluate the regional similarity between the predicted segmentation and the ground-truth in binary setting for end-to-end optimization. Finally, we employ majority voting to  fuse all the binary outputs within the same slice and concatenate all fused slices to generate the volumetric multi-organ mask as our refined output.

\section{Experimental Setup}
\subsection{Datasets}
\textbf{[I] MICCAI 2015 BTCV Challenge} is comprised of 100 de-identified 3D contrast-enhanced CT scans with 7968 axial slices. 20 available scans are publicly available for testing phase. This dataset includes 12 well-annotated organs, including the spleen, right kidney, left kidney, gall bladder, esophagus, liver, stomach, aorta, inferior vena cava (IVC), portal splenic vein (PSV), pancreas and right adrenal gland. Each volume consists of $47\sim133$ slices of $512\times512$ pixels at a resolution of with resolution of $([0.54\sim0.98]\times[0.54\sim0.98]\times[2.5\sim7.0])mm^{3}$.\par 
\textbf{[II] Research Non-Contrast CT cohorts} is retrieved and consists of 56 volumetric CT scans with 3687 axial slices and expert refined annotations for the same 12 organs in BTCV dataset. Each volume consists of $49\sim174$ slices of $512\times512$ pixels, with resolution of $([0.64\sim0.98]\times[0.64\sim0.98]\times[1.5\sim5.0])mm^{3}$. \par 
\textbf{[III] MICCAI 2021 FLARE Challenge \cite{ma2021abdomenct}} adapts large scale of abdominal contrast-enhanced CT with 511 cases from 11 medical centers. In total, 361 CT consists of well-annotated labels for spleen, kidney, liver and pancreas organs. Each volume consists of $43\sim384$ slices of $512\times512$ pixels, with resolution of $([0.64\sim0.98]\times[0.64\sim0.98]\times[1.0\sim5.0])mm^{3}$.

\begin{figure*}
\centering
\includegraphics[width=\textwidth]{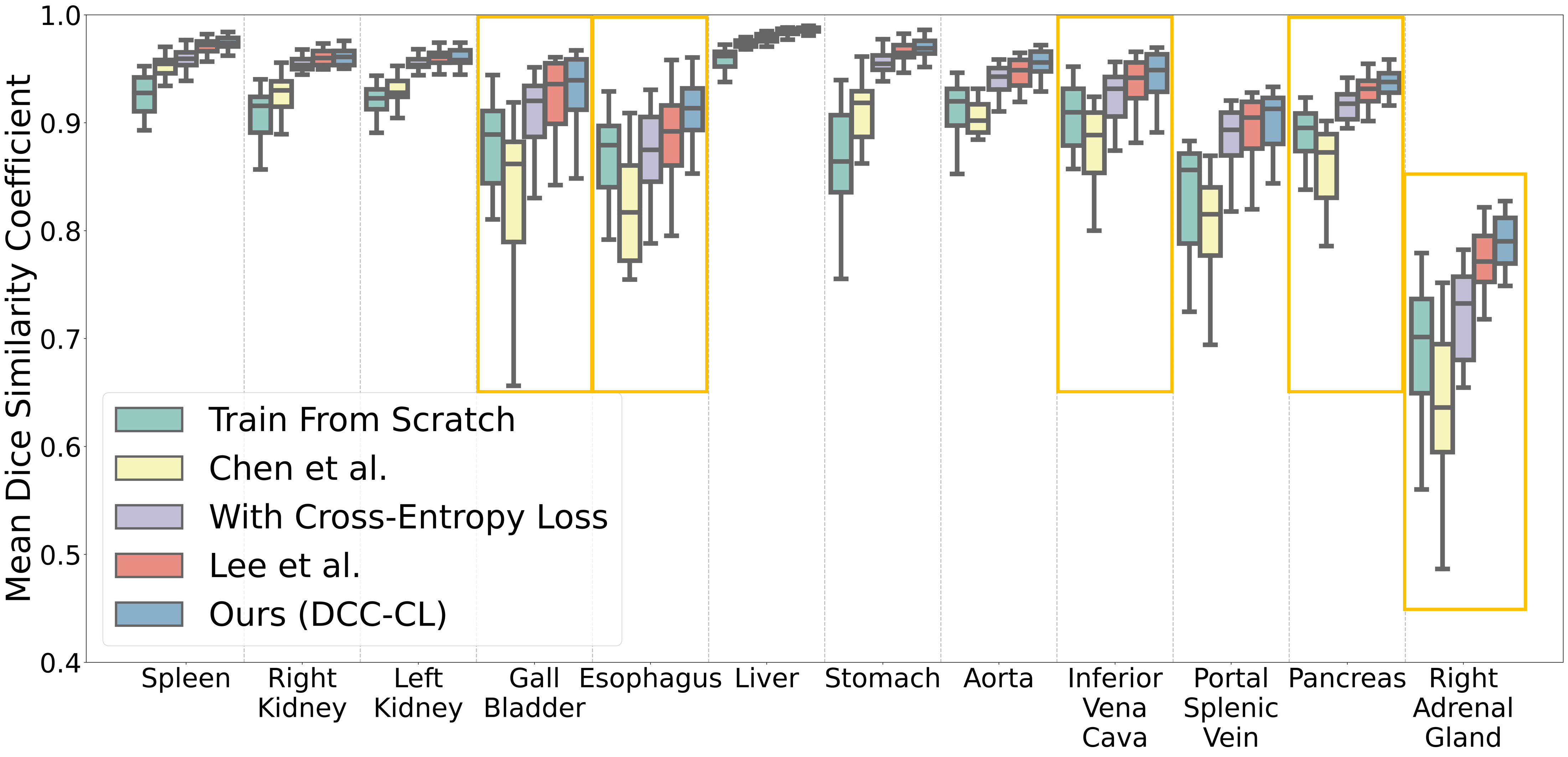}
\caption{DCC-CL outperforms the current state-of-the-art contrastive learning methods with $Chen\:et\:al.$ and $Lee\:et\:al.$. The yellow box demonstrates the significant improvement in organs with subtle contrast variation between modalities.} \label{qual_figure}
\end{figure*}

\subsection{Data Preprocessing}
Before we input the complete volume into the network for coarse multi-organ segmentation, we apply hierarchical steps for data preprocessing. The first step is to perform soft tissue windowing is applied between the range of -175 and 250 Hu. Intensity normalization is performed after windowing for each volume and use min-max normalization: $(X-X_1)/(X_{99}-X_1)$ to normalize the intensity value between 0 and 1, where $X_p$ denote as the $p^{th}$ intensity percentile in $X$. After intensity normalization, body part regression algorithm is applied to each volume and extract the field of view for abdominal region only for downstream segmentation. The regression model define value from -12 to +12 as the whole body. We define the anatomical value between -4 to 5 as the abdominal region for cropping.

\subsection{Model \& Training Details}
For contrastive pretraining step, the complete backbone of the network consists of a ResNet-50 image encoder and a projection network with two linear layers. For organ segmentation finetuning step, DeepLabV3+ network is employed by withdrawing the projection network and following with a decoder using ASPP modules. Both pretraining and finetuning process are optimized with an Adam optimizer (weight decay: $1\times10^-4$; momentum: 0.9; batch size: 4). We pretrain the model with our contrastive approach for 10 epochs using a learning rate of 0.0003, while we finetune the model with 5 epochs with learning 0.0001. On a NVIDIA-Quadro RTX 5000, 1 epoch of pretraining takes about 12 hours to finish with batch size of 4.

\subsection{Implementation Details}
For internal evaluation, we evaluate the model performance with a downstream multi-organ segmentation task and compare with current state-of-the-art of fully-supervised and contrastive learning approaches for both BTCV and the in-house non-contrast CT datasets. For external validation, we randomly selected 100 samples from FLARE dataset and perform inference with the optimized model trained with internal cohorts. We use Dice similarity coefficient as an evaluation metric to compare the overlapping regions between predictions and ground-truth label. Furthermore, we performed ablation studies with hyperparameter variation and different training strategies to pretrain the encoder network. 

\begin{table*}[t!]
    \centering
    \caption{Ablation studies of segmentation performance with different pretraining scenarios of the BTCV testing cohort.}
    \resizebox{2\columnwidth}{!}{%
    \begin{tabular}{{c c|}*{13}{c}}
        \toprule\toprule
        Encoder & Pretrain & Spleen & R.Kid & L.Kid & Gall. & Eso. & Liver & Stomach & Aorta & IVC & PSV & Pancreas & R.A & Mean\\
        \midrule
        ResNet50 & $\times$ & 0.932 & 0.877 & 0.887 & 0.860 & 0.761 & 0.962 & 0.941 & 0.832 & 0.815 & 0.735 & 0.833 & 0.587 & 0.840\\
        ResNet50 & SSCL & 0.953 & 0.922 & 0.930 & 0.842 & 0.822 & 0.972 & 0.907 & 0.899 & 0.874 & 0.800 & 0.854 & 0.625  & 0.868 \\
        ResNet50 & CE & 0.959 & 0.948 & 0.957 & 0.890 & 0.868 & 0.978 & 0.956 & 0.935 & 0.919 & 0.884 & 0.903 & 0.725  & 0.905 \\
        ResNet50 & AGCL & 0.971 & 0.955 & \textbf{0.963} & 0.910 & 0.886 & 0.984 & 0.965 & 0.941 & 0.932 & 0.893 & 0.917 & 0.769 & 0.923 \\
        ResNet50 & DCC-CL & \textbf{0.974} & \textbf{0.956} & 0.960 & \textbf{0.928} & \textbf{0.905} & \textbf{0.986} & \textbf{0.968} & \textbf{0.950} & \textbf{0.939} & \textbf{0.901} & \textbf{0.923} & \textbf{0.776} & \textbf{0.936}\\
        \bottomrule\bottomrule
    \end{tabular}%
    }
    \label{ablation_resnet}
\end{table*}

\begin{table*}[t!]
    \centering
    \caption{Ablation studies of segmentation performance in various network backbones of the in-house non-contrast testing cohort.}
    \resizebox{2\columnwidth}{!}{%
    \begin{tabular}{{c c|}*{13}{c}}
        \toprule\toprule
        Encoder & Pretrain & Spleen & R.Kid & L.Kid & Gall. & Eso. & Liver & Stomach & Aorta & IVC & PSV & Pancreas & R.A & Mean\\
        \midrule
        ResNet50 & $\times$ & 0.960 & 0.918 & 0.921 & 0.754 & 0.783 & 0.964 & 0.950 & 0.840 & 0.839 & 0.691 & 0.796 & 0.372 & 0.816\\
        ResNet50 & SimCLR & 0.964 & 0.938 & 0.946 & 0.800 & 0.801 & 0.969 & 0.946 & 0.901 & 0.869 & 0.739 & 0.804 & 0.386 & 0.848\\
        ResNet50 & CE & 0.972 & 0.952 & 0.961 & 0.812 & 0.859 & 0.974 & 0.959 & 0.934 & 0.914 & 0.768 & 0.838 & 0.551 & 0.875\\
        ResNet50 & AGCL & 0.982 & 0.962 & 0.965 & 0.834 & 0.879 & 0.982 & 0.967 & 0.945 & 0.929 & 0.790 & 0.850 & 0.560 & 0.892\\
        ResNet50 & DCC-CL & \textbf{0.984} & \textbf{0.966} & \textbf{0.970} & \textbf{0.957} & \textbf{0.890} & \textbf{0.982} & \textbf{0.971} & \textbf{0.951} & \textbf{0.939} & \textbf{0.803} & \textbf{0.893} & \textbf{0.584} & \textbf{0.910}\\
        \bottomrule\bottomrule
    \end{tabular}%
    }
    \label{tab:my_label}
\end{table*}

\begin{figure*}[t!]
    \centering
    \begin{subfigure}[b]{0.49\textwidth}
    \includegraphics[width=\textwidth]{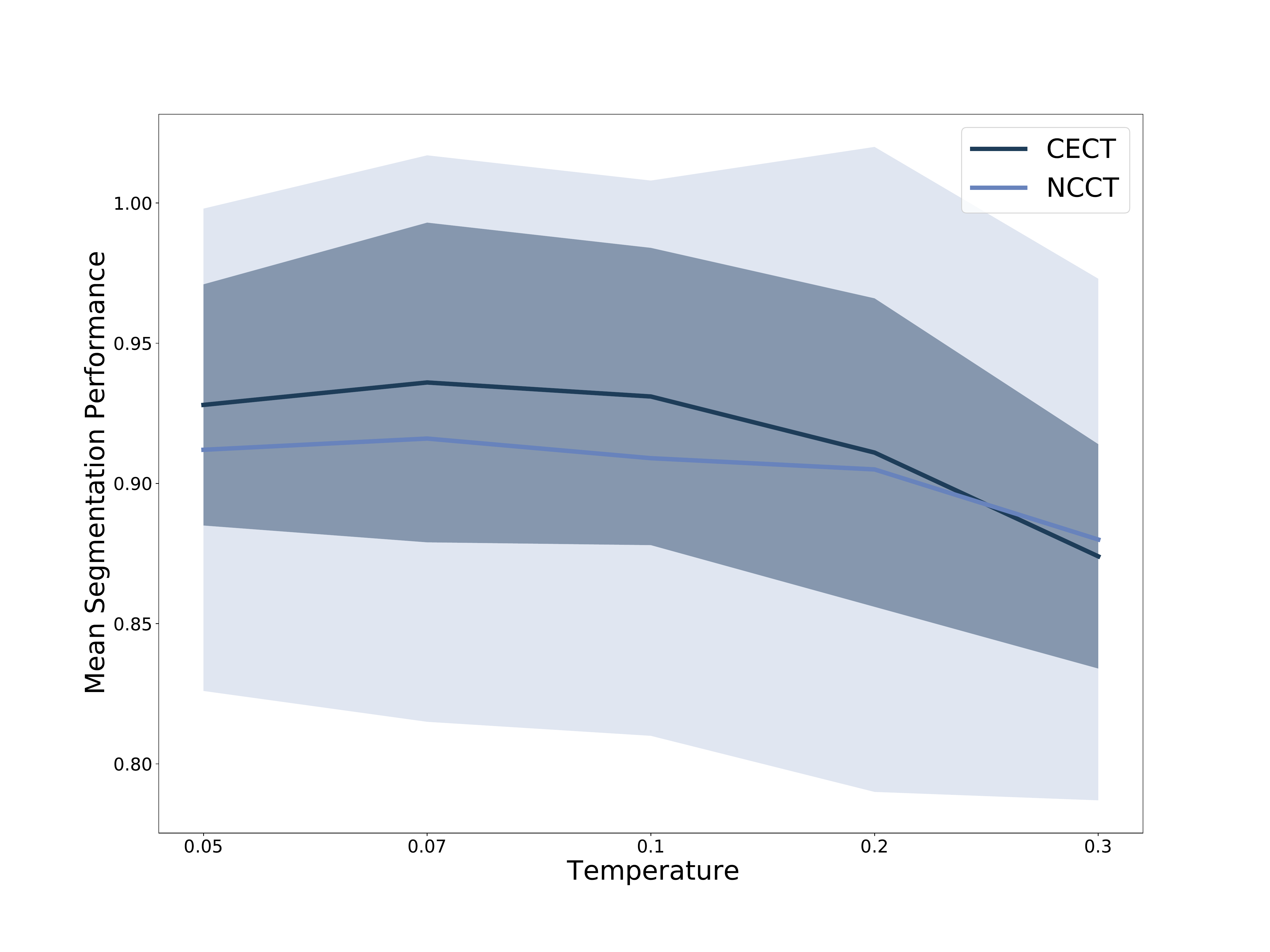}
    \caption{}
    \end{subfigure}
    \begin{subfigure}[b]{0.49\textwidth}
    \includegraphics[width=\textwidth]{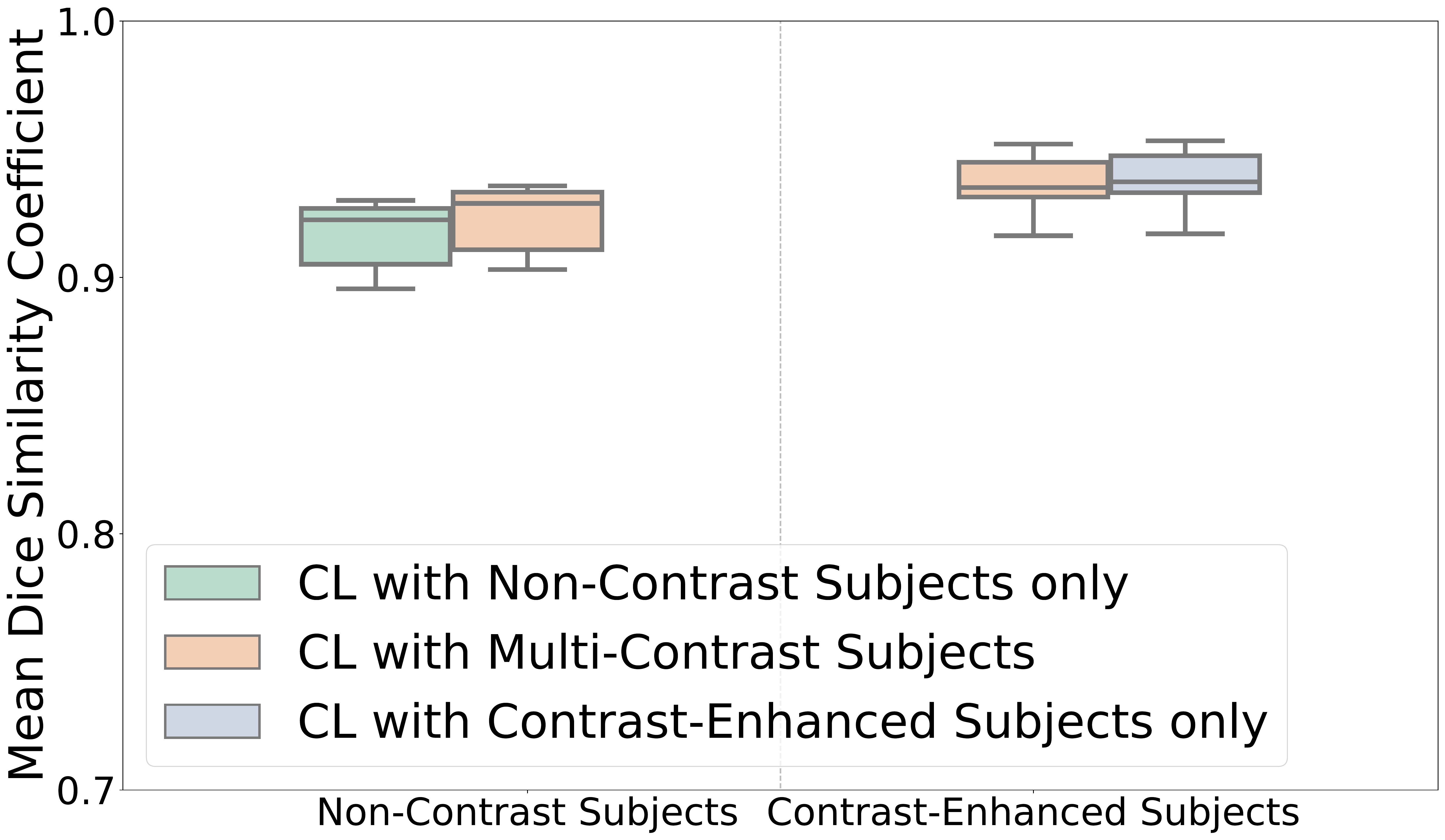}
    \caption{}
    \end{subfigure}
\caption{\textbf{a)} Variability of temperature scaling demonstrates that the segmentation performance is best optimized when $T=0.07$. \textbf{b)} Segmentation performance of NCCT is significantly improved without trading off CECT performance.}
\end{figure*}

\section{Experimental Results}
\subsection{Comparison with Fully/Partially Supervised Approaches} 
As the preliminary basis of our approach is to extract organ-specific patches with a coarse segmentation network, we evaluate and validate current supervised state-of-the-arts as our coarse segmentation backbone. All supervised approaches are presented in Table 1 and 2 for both internal cohorts, including 3D CNN-based network \cite{cciccek20163d}, hierarchical networks \cite{roth2018multi, zhu2019multi, lee2021rap} and transformer-based network \cite{hatamizadeh2022unetr}. Significant levels of improvement (from mean Dice 0.784 to 0.836, $p>0.01$) are demonstrated starting from \cite{cciccek20163d} to \cite{hatamizadeh2022unetr} for both contrast-enhanced BTCV and in-house non-contrast cohorts. \cite{lee2021rap} achieves the best segmentation performance in BTCV, while \cite{hatamizadeh2022unetr} demonstrates the best segmentation performance in non-contrast dataset. As the model is trained coherently in multi-contrast setting, the trained model may bias to one of the contrast modality and thus lead to a level of degradation on certain organ segmentation performance with specific contrast. Furthermore, \cite{Zhou_2019_ICCV} demonstrates to leverage partial labeled samples to enhance the performance for multi-organ segmentation. We convert the partial supervised training scenarios of \cite{Zhou_2019_ICCV} as using one contrast samples for downstream multi-organ segmentation, while the another contrast samples are used in partially labeled setting. Interestingly, \cite{Zhou_2019_ICCV} demonstrates a stable improvement on both BTCV (mean Dice score from 0.842 to 0.850) and in-house non-contrast (mean Dice score from 0.815 to 0.833) cohorts. Such improvement in performance may due to the organ-specific prior distribution generated from \cite{Zhou_2019_ICCV} is beneficial to adapt the contrast variation in multi-organ segmentation. By leveraging the psuedo output from the supervised-trained model as organ attentional guidance for contrastive learning, the downstream segmentation performance is significantly enhanced across the contrastive learning state-of-the-arts.

\subsection{Comparison with Contrastive Learning Approaches} 
After looking into the supervised approahces, we compare our proposed DCC-CL with a series of contrastive learning approaches for both internal testing cohorts, including SimCLR (\cite{chen2020simple}), local and global contrastive learning (\cite{chaitanya2020contrastive}), DenseCon (\cite{wang2021dense}), SupCon (\cite{khosla2020supervised}), PixelCon \cite{wang2021exploring} and multi-label contrastive learning (\cite{hin2021semantic}). By adapting self-supervised approaches with the organ-attention guidance, we adapt the intrinsic structure of SimCLR to define the organ-attentional representation in the latent space and demonstrates an significant improvement of 2.49\% and 6.27\% of Dice score for BTCV and non-contrast cohorts respectively. By further extract representations in both local and global field of view, a slightly increase is demonstrated from Dice score 0.863 to 0.874 for BTCV and 0.848 to 0.854 for non-contrast cohort. Apart from using linear projection to extract global-wise features, DenseCon propose to adapt dense projection and evaluate the similarity between each feature vector in the dense mappings. Interestingly, a significant enhancement with 3.86\% Dice score ($p>0.01$) is demonstrated for non-contrast cohort, while only a slight increase with 0.915\% Dice score is shown for BTCV.

\begin{table}[!t]
    \centering
    \caption{Comparison of the current state-of-the-art methods on FLARE dataset.}
    \resizebox{\columnwidth}{!}{%
    \begin{tabular}{*{1}{l}|*{4}{c}|*{1}{c}}
        Method & Spleen & Kidney & Liver & Pancreas &\tabincell{c}{Average Dice} \\
        \toprule
        \cite{lee2021rap} & 0.956 & 0.903 & 0.954 & 0.730 & 0.885\\
        \cite{chen2020simple} & 0.960 & 0.910 & 0.960 & 0.756 & 0.896\\
        \cite{chaitanya2020contrastive} & 0.961 & 0.923 & 0.956 & 0.787 & 0.908\\
        \cite{wang2021dense} & 0.966 & 0.918 & 0.964 & 0.800 & 0.912\\
        \cite{khosla2020supervised} & 0.963 & 0.918 & 0.966 & 0.830 & 0.919\\
        \cite{wang2021exploring} & 0.968 & 0.940 & 0.964 & 0.811 & 0.922\\
        \cite{hin2021semantic} & 0.971 & \textbf{0.948} & 0.968 & 0.823 & 0.927\\
        \midrule
        \textbf{Ours (DCC-CL)} & \textbf{0.975} & 0.944 & \textbf{0.971} & \textbf{0.847} & \textbf{0.934}\\
    \end{tabular}%
    }
    \label{tab:my_label}
\end{table}

\cite{khosla2020supervised} and \cite{hin2021semantic} further extend the self-supervised contrastive loss and leverage single/multi-class labels to define conditional positive pairs for learning independent embedding of each semantic class. The integration of image-level labels demonstrates significant improvement on finetuning downstream segmentation task. \cite{wang2021exploring} extract the semantic context from the segmentation and define the class-wise positive pairs from the psuedo predictions. Interestingly, both \cite{khosla2020supervised} and \cite{wang2021exploring} demonstrate significant improvement of Dice score 0.907 and 0.913 for BTCV respectively, while slightly decreases of performance are shown in the non-contrast cohort. The reason for such degradation may correspond to the contrast variation and the learned representations only defined as class-wise embeddings without contrast information. \cite{hin2021semantic} target the limitation of using contrast information and adapt both contrast and organ label to define representations in the latent space. \cite{hin2021semantic} outperforms the previous contrastive learning state-of-the-arts for both internal cohorts with Dice score of 0.923 and 0.892 respectively. By further adapting the contrast correlation as dynamic weighting constraints, DCC-CL achieves the best performance among all state-of-the-arts with a mean Dice score of $0.936$ and $0.910$ for both BTCV and non-contrast cohorts respectively. The additional gains demonstrate that the use of data-driven information provides flexibility to control and define multi-phase latent spaces. 

Furthermore, we perform external evaluations on FLARE dataset with all contrastive state-of-the-art approaches, as shown in Table 3. As we leverage RAP-Net (\cite{lee2021rap}) as the coarse segmentation backbone, it demonstrates a stable performance on external dataset with a mean dice of 0.885. A stable enhancement of performance is demonstrated across all state-of-the-arts from Dice score of 0.885 to 0.927. Leveraging the dynamic weighting with contrast correlations further benefit the generalizability of the model and shows that DCC-CL outperforms all contrastive learning state-of-the-arts with the mean Dice of 0.927 to 0.934 . \par

\subsection{Ablation Studies} 
\textbf{Comparing with Contrastive Pretraining Strategies.}
To investigate the effect of using image-level contrast correlation as dynamic weight for contrastive learning, we perform multiple pretraining scenarios to compare the downstream tasks performance: 1) training from scratch, 2) pretraining with self-supervised contrastive loss (SSCL, \cite{chen2020simple}), 3) pretraining with cross-entorpy (CE) loss as classification tasks, 4) pretraining with multi-label contrastive loss (AGCL, \cite{hin2021semantic}) in ResNet-50 encoder backbone. We start with the scenario of "training from scratch" and leverage the self-supervised contrastive loss for pretraining as the basis of comparison. After that, instead of using contrastive loss to constrain the representations, we further implemented the classification scenario by classifying the representations into organ and contrast via the canonical CE loss. We finally compare the scenario of leveraging hard multi-class label with our proposed scenario. All pretraining scenarios are to define representations into corresponding contrast and organ defined embeddings. 

As shown in Fig. 4 and Table 3, pretraining with SSCL demonstrates a significant improvement of 3.12\% Dice than the scenario of without pretraining. Such improvement is expected that pretraining with SSCL provides an initial definition of the learned representations in the latent space and hypothesis that such learned embeddings are beneficial to the downstream tasks. The scenario of without pretraining is only relied on the ability of the decoder and define representations with the downstream tasks guidance. By adapting the hard image-level label for pretraining, the segmentation performance significantly boost from Dice score 0.868 to 0.905 and the performance further enhances to 0.923 by leveraging the multi-label into the contrastive loss. Using CE loss as pretraining is to classify representations into class-wise embeddings with the labels given and the learned representations in the same class are moved towards each other. By adapting the multi-label context into contrastive loss, AGCL demonstrates the flexibility of move the same class representation near and push the unrelated classes representations away in the latent space, improving the segmentation performance with a better definition of latent space. Eventually, DCC-CL surpasses AGCL by 1.41\% in mean Dice. Instead of providing hard labels for searching similar representations, the computed contrast correlation matrix can leverage as a soft weighting to control the cosine similarity between pairwise representations and adapt the contrast variation more effectively across organ of interests.  

\textbf{Temperature Variability.}
We further vary temperature to investigate the effect of temperature on finetuning the segmentation task across scales. Fig. 4a demonstrates the effect of temperature across all subjects in BTCV testing dataset. We find that the optimal temperature is roughly around 0.07 and the segmentation performance starts to degrade with the increasing temperature. Low temperature tends to penalize more to the highly similar representations and enhance the difficulties of searching positive pairs. Therefore, computing the contrastive loss with low temperature may achieve better segmentation performance that that using high temperature. 

\textbf{Single/Multiple Phase Contrastive Learning.}
In the training scenario, we also want to investigate the effectiveness of applying DCC-CL with single phase CT only and with multi-phases CT for downstream segmentation. In Fig. 4b, we observed that the segmentation performance with non-contrast cohort is significantly improved by training in multi-contrast scenario, while the segmentation performance with BTCV is comparatively similar to each other. The contrast correlation matrix aims to generalize different level of contrast variation and the segmentation performance with both modalities samples are well preserved without trading off one of the modalities performances.

\subsection{Discussion \& Limitations} 
In this study, we present an adaptive contrastive pre-training framework that leverages the variable level of contrast characteristics in each organ of interests to guide the separability of semantic embeddings. We hypothesize that such dynamically defined latent space is more beneficial to the downstream segmentation task, compared to the hard label defined latent space. With the organ attention guidance, the model allows to extract meaningful representations within the organ attention region. Also, we can further extract additional context with signficant variability from the image itself (e.g. contrast). Instead of using one-hot hard labels to define the "scan-wise" contrast level,  the similarity between the mean intensity demonstrates to be a dynamic soft constrain and define the representations adaptively. From Fig. 3, the PCA plot demonstrates the separability of the corresponding organ-wise representations. The orange bounding box localizes the organ representations with similar contrast, while the images are sampled from different contrast phases. The corresponding image visualization further demonstrates the contrast correspondence in the organ and the effectiveness of leveraging the contrast information for pretraining. In both Table 1 and 2, we have shown that our proposed pretraining strategy outperforms the current contrastive state-of-the-art approaches for multi-contrast organ segmentation. Meanwhile, we preseve the organ segmentation performance by training in the multi-contrast setting, instead of trading off the performance from one of the modalities. Such dynamic weighting further provides a better understanding on leveraging soft labels to define latent space for downstream segmentation task. 

One limitation of DCC-CL is its dependence on coarse pseudo labels. As the 2D patches are extracted with the pseudo label guidance, inaccurate organ patches may also be extracted and be unable to provide accurate organ-wise intensity information. Another limitation is that the representations are learned in an organ-centric setting. We aim to adapt the contrast correlation between the neighboring organs to an end-to-end framework in our future work.

\section{Conclusion}
Adapting the multi-phase contrast imaging with deep learning models remains a persistent challenge for performing robust semantic segmentation with wide range of contrast variation. In this work, we propose a novel contrastive loss function that control the distance between representations with dynamic contrast correlation guidance. Our proposed contrastive loss contributes a significant gain of the segmentation performance across multi-phase contrast CT.\\

\section*{Acknowledgments}
This research is supported by NIH Common Fund and National Institute of Diabetes, Digestive and Kidney Diseases U54DK120058 (Spraggins), NSF CAREER 1452485, NIH 2R01EB006136, NIH 1R01EB017230 (Landman), and NIH R01NS09529. This study was in part using the resources of the Advanced Computing Center for Research and Education (ACCRE) at Vanderbilt University, Nashville, TN. The identified datasets used for the analysis described were obtained from the Research Derivative (RD), database of clinical and related data. The imaging dataset(s) used for the analysis described were obtained from ImageVU, a research repository of medical imaging data and image-related metadata. ImageVU and RD are supported by the VICTR CTSA award (ULTR000445 from NCATS/NIH) and Vanderbilt University Medical Center institutional funding. ImageVU pilot work was also funded by PCORI (contract CDRN-1306-04869).

\bibliographystyle{model2-names.bst}\biboptions{authoryear}
\bibliography{medima-template.bib}



\end{document}